\documentclass{article}

\usepackage{arxiv}

\usepackage[utf8]{inputenc} 
\usepackage[T1]{fontenc}    
\usepackage{hyperref}       
\usepackage{url}            
\usepackage{booktabs}       
\usepackage{amsfonts}       
\usepackage{nicefrac}       
\usepackage{microtype}      
\usepackage[font=small,labelfont=small,skip=1.5\baselineskip]{caption}
\usepackage{longtable}
\usepackage{graphicx}
\usepackage{doi}

\title{Deep learning approaches to Earth observation change detection}

\author{Antonio Di Pilato\\
	Dipartimento Interateneo di Fisica\\
	Università degli Studi di Bari\\
    Bari, Italy\\
	\texttt{antonio.dipilato@ba.infn.it}\\
	\And
	Nicolò Taggio\\
	Planetek Italia\\
	Bari, Italy\\
	\texttt{taggio@planetek.it}\\
	\And
	Alexis Pompili \\
	Dipartimento Interateneo di Fisica\\
	Università degli Studi di Bari\\
    Bari, Italy\\
	\texttt{alexis.pompili@ba.infn.it} \\
	\And
	Michele Iacobellis\\
	Planetek Italia\\
	Bari, Italy\\
	\texttt{iacobellis@planetek.it}\\
    \And
    Adriano Di Florio \\
	Dipartimento Interateneo di Fisica\\
	Politecnico di Bari\\
    Bari, Italy\\
	\texttt{adriano.diflorio@ba.infn.it} \\    
	\And
	Davide Passarelli\\
	Planetek Italia\\
	Bari, Italy\\
	\texttt{passarelli@planetek.it} \\
	\And
	Sergio Samarelli\\
	Planetek Italia\\
	Bari, Italy\\
	\texttt{samarelli@planetek.it}

}

\date{}

\renewcommand{\shorttitle}{\textit{arXiv} Template}

\hypersetup{
pdftitle={A template for the arxiv style},
pdfsubject={q-bio.NC, q-bio.QM},
pdfauthor={David S.~Hippocampus, Elias D.~Striatum},
pdfkeywords={First keyword, Second keyword, More},
}

\begin{document}
\maketitle

\begin{abstract}
	The interest for change detection in the field of remote sensing has increased in the last few years. Searching for changes in satellite images has many useful applications, ranging from land cover and land use analysis to anomaly detection. In particular, urban change detection provides an efficient tool to study urban spread and growth through several years of observation. At the same time, change detection is often a computationally challenging and time-consuming task, which requires innovative methods to guarantee optimal results with unquestionable value and within reasonable time. In this paper we present two different approaches to change detection (semantic segmentation and classification) that both exploit convolutional neural networks to achieve good results, which can be further refined and used in a post-processing workflow for a large variety of applications.
\end{abstract}


\section{Introduction}
Change detection in the field of remote sensing is the process of identifying differences in the state of an object or phenomenon by observing it at different subsequent times~\cite{cdtask}. According to this definition, such task involves the usage of methods and algorithms that compare two or more satellite images of the same scene and produce results that can either map changes pixel by pixel or provide a score indicating the presence of more or less relevant changes within the area.

Specifically, change detection applied to urban areas~\cite{urbancd1, urbancd2} is an interesting case study, as it aims at monitoring differences in land cover due to urban expansion and spread over the years in a more automated way. Although a large variety of methods exists and has been classified and discussed~\cite{cdmethods}, the deep learning approach to this problem is currently object of active research.

At the same time, training models with a high level of generalization capabilities is extremely challenging, due to the changing urban patterns from a country to another. In addition, vegetation and seasonal changes often represent a source of contamination that cannot be easily removed, and contribute to lower the performance of efficient convolutional neural networks (CNNs).

In this scenario, performing change detection with CNNs has the objective of providing a fast and useful analysis of an area that can be integrated into traditional post-processing workflows on the ground stations, which combine results from several algorithms to obtain a final map of all the changes present in the scene. This objective can be achieved with two different approaches: semantic segmentation and classification. The first approach is more suitable for the full analysis on ground: the acquired data are downloaded at the ground facilities and the deep learning algorithm provides a detailed map of all the changes detected within the area. The second approach is onboard-oriented, as it aims to evaluate the change content of a scene and transmit such estimation to the ground stations, which can further decide whether to download the new acquired data for additional analyses.
Specific monitoring tasks can be planned exploiting the onboard processing to filter only the interesting data, thus mitigating the download bottleneck. 

\section{The Sentinel-2 mission}
The Sentinel-2 mission consists of a constellation of two twin satellites flying in the same Sun-synchronous polar orbit but phased at $180^{\circ}$ at a mean altitude of $786$~km, designed to give a high revisit frequency of five days at the Equator. It aims at monitoring the variability of the land surface conditions through the acquisition of high resolution multispectral images that can be exploited for land cover/change classification, atmospheric correction and cloud/snow separation.

Indeed, each satellite carries a multispectral instrument (MSI) for data acquisition~\cite{s2msi}. The MSI measures the Earth's reflected radiance in $13$ spectral bands: the visible and near-infrared (VNIR), and the short-wave infrared (SWIR), with three different spatial resolutions ($10$~m, $20$~m and $60$~m). Additional information is provided in Table~\ref{tab:s2bands}.

\begin{table}[t!]
    \centering
    \begin{tabular}{|c|c|c|c|c|c|}
    \hline
    \multicolumn{1}{|l|}{} & \multicolumn{2}{c|}{S2A} & \multicolumn{2}{c|}{S2B} & \multicolumn{1}{l|}{} \\ \hline
    \begin{tabular}[c]{@{}c@{}}Band\\ Number\end{tabular} & \begin{tabular}[c]{@{}c@{}}Central\\ wavelength \\ (nm)\end{tabular} & \begin{tabular}[c]{@{}c@{}}Bandwidth\\ (nm)\end{tabular} & \begin{tabular}[c]{@{}c@{}}Central\\ wavelength\\ (nm)\end{tabular} & \begin{tabular}[c]{@{}c@{}}Bandwidth\\ (nm)\end{tabular} & \begin{tabular}[c]{@{}c@{}}Spatial\\ resolution\\ (m)\end{tabular} \\ \hline
    1 & 442.7 & 21 & 442.3 & 21 & 60 \\
    2 & 492.4 & 66 & 492.1 & 66 & 10 \\
    3 & 559.8 & 36 & 559.0 & 36 & 10 \\
    4 & 664.6 & 31 & 665.0 & 31 & 10 \\
    5 & 704.1 & 15 & 703.8 & 16 & 20 \\
    6 & 740.5 & 15 & 739.1 & 15 & 20 \\
    7 & 782.8 & 20 & 779.7 & 20 & 20 \\
    8 & 832.8 & 106 & 833.0 & 106 & 10 \\
    8a & 864.7 & 21 & 864.0 & 22 & 20 \\
    9 & 945.1 & 20 & 943.2 & 21 & 60 \\
    10 & 1373.5 & 31 & 1376.9 & 30 & 60 \\
    11 & 1613.7 & 91 & 1610.4 & 94 & 20 \\
    12 & 2202.4 & 175 & 2185.7 & 185 & 20 \\ \hline
    \end{tabular}
    \vspace{3mm}
    \caption{Spectral bands of the Sentinel-2 twin satellites' sensors~\cite{s2msi}.}
    \label{tab:s2bands}
\end{table}

Sentinel-2 data are downloaded at the ground segment, which uses several processing algorithms to obtain two types of final products, in the form of compilations of elementary granules or tiles (minimum indivisible partitions) of fixed size acquired within a single orbit. The Level-1C (L1C) products are composed by $100\times100$~km$^2$ tiles, ortho-images in UTM/WGS84 projection. Pixel values in L1C products are provided as top-of-atmosphere (TOA) reflectance measurements, along with the parameters to transform them into radiance measurements. Cloud masks and information about ozone, water vapour and mean sea level pressure are also included. The Level-2A (L2A) products consist of bottom-of-atmosphere (BOA) reflectance images derived from the associate L1C products by means of atmospheric correction algorithms. Therefore, each product is composed by $100\times100$~km$^2$ tiles in cartographic geometry (UTM/WGS84 projection). L2A products are generated at the ground segment since $2018$, but the L2A processing method can be applied by users to even older data through the Sentinel-2 Toolbox.

\subsection{Change detection dataset}
A common limitation of using deep learning algorithms for change detection is the poor availability of already-labeled datasets. Despite large amounts of satellite data are acquired and downloaded daily, the process of assigning a ground truth to images in any format (depending on the specific task) is a well-known time-consuming operation. In this study, the Onera Satellite Change Detection (OSCD) dataset~\cite{onera} was used for the purpose. It consists of $24$ Sentinel-2 L1C image pairs, each capturing an area of approximately $600 \times 600$ pixels at $10$~m resolution; bands at $20$~m and $60$~m resolution of the L1C products were upsampled to the resolution of $10$~m such that all the channels have aligned pixels. Even though the ground truths of all the $24$ image pairs are currently available, only $14$ regions were used for training purpose, as originally suggested by the authors of the dataset. In addition, ground truths are provided in the form of binary change maps where each pixel has a value marking if a change occurred at that location within the original image pair or not.

\begin{figure}[tb]
	\centering
    \includegraphics[width=.3\textwidth]{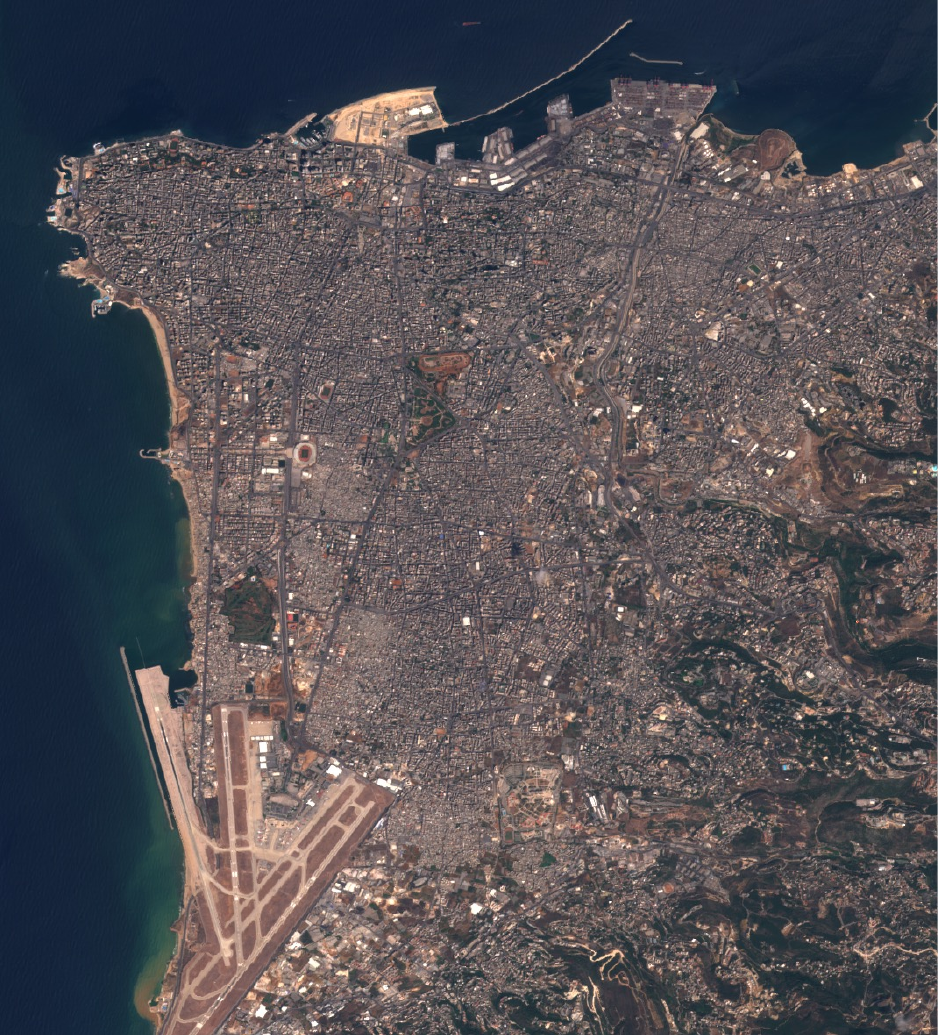}
    \hspace{0.3cm}
    \includegraphics[width=.3\textwidth]{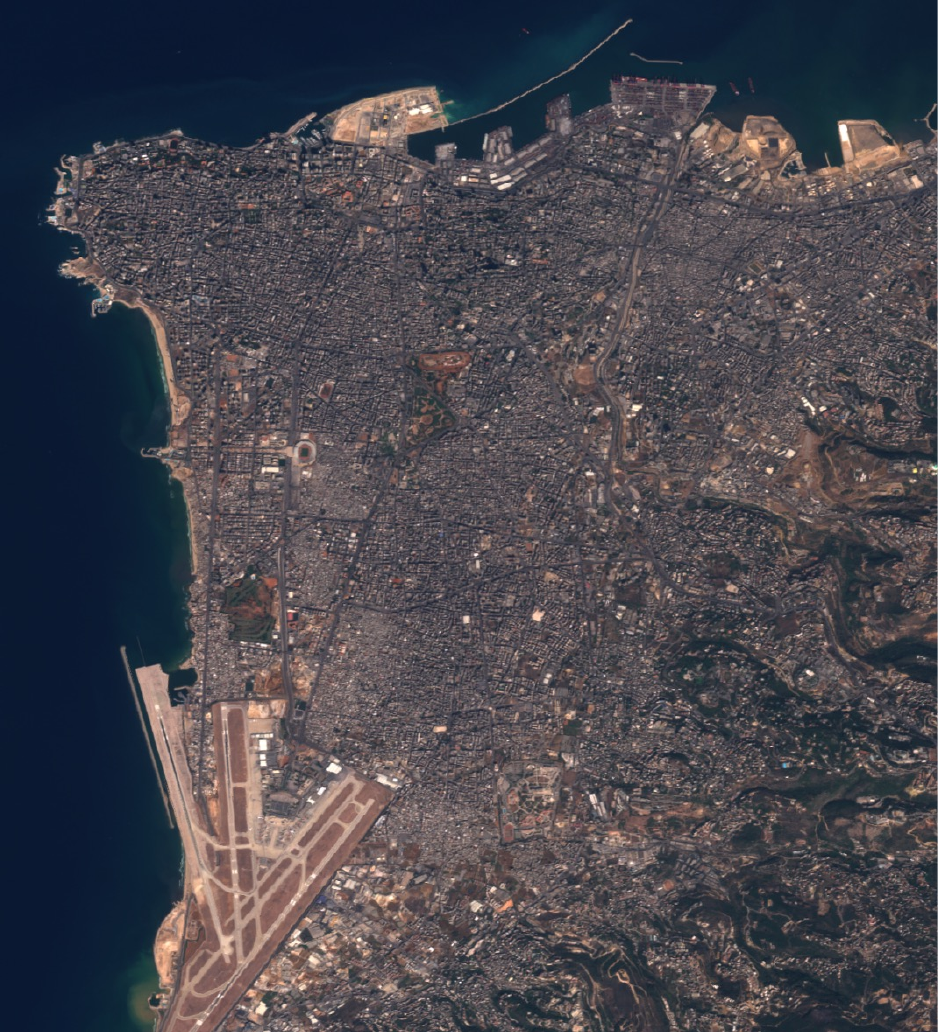}
    \hspace{0.3cm}
    \includegraphics[width=.3\textwidth]{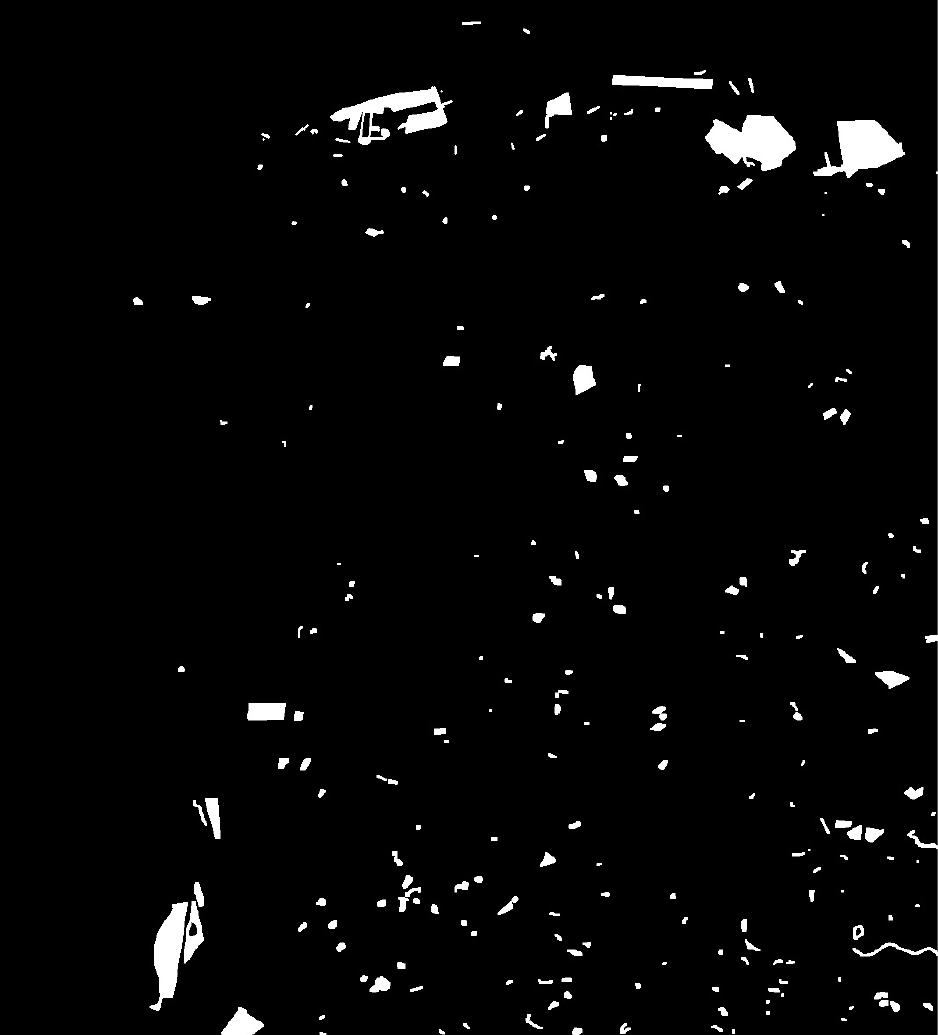}
	\caption{Sample image pair and corresponding ground truth of the Beirut area. Left is the image before changes (20 August 2015), center is the image after changes occurred (3 January 2017) and right is the change map. White pixels map the changes detected in the scene~\cite{onera}.}
	\label{fig:beirut_sample}
\end{figure}

Images of each pair were taken at a temporal distance of $2$-$3$ years; a very small amount of clouds is present in some of them and no restrictions were set on brightness conditions, as vegetation, seasonal and sunlight exposure changes were ignored. A sample image pair (RGB-only) and its corresponding change map are shown in Figure~\ref{fig:beirut_sample}.

\subsection{Dataset splitting strategy}
In order to have a rich training dataset for the purpose of the study, patches with size of $128 \times 128 \times 13$ were extracted from the original $14$ image pairs according to the following method. First, a random pixel location $(x,y)$ was selected within the original pre-changes image and the area of pixels $[x,y,x+a,y+a]$, with $a=128$, was cropped. Such selection was operated while having the cropped patch entirely confined in the scene borders. The same crop was extracted from the corresponding post-changes image and the change map. Then, the ratio $R$ between the number of changed pixels and the total number of pixels for the selected patch was evaluated on the binary change map and two possible cases established:
\begin{enumerate}
\itemsep0em
    \item $0\leq R<0.1$ (changes cover less than $10\%$ of the area captured by the patch);
    \item $R\geq0.1$ (the $10\%$ or more of the area captured by the patch is covered by change pixels).
\end{enumerate}
In the first case, a random transformation among six possible choices (rotation by $90^{\circ}$, $180^{\circ}$, $270^{\circ}$ or $360^{\circ}$, vertical or horizontal flip) was applied to the two images of the pair and to the corresponding change map, and the results were stored in the training dataset. In the second case, instead, all the six available transformations were operated, one per time, and the resulting patches and change maps saved and stored as well. A total of $500$~crops were randomly selected for each of the $14$ regions; eventual duplicates were removed and the final training dataset consisted of $13919$~input pairs and ground truths.

It should be observed that the randomness introduced through the crop selection and the applied transformation reduces the effect of pixel correlation due to patches being extracted from the same original region. Indeed, patches do not share a fixed number of pixels which might heavily bias the optimization step during the training process. Furthermore, those patches that share a certain amount of pixels can be still distinguished by a transformation, enhancing such reduction effect.

Finally, while for the semantic segmentation approach ground truths are originally available in the OSCD dataset, for the classification task they were generated by applying the following criterion: if the changed pixels of the $128\times128$ ground truth are less than $25$, then the image pair is labeled as $0$ (not interesting for further analysis on ground); otherwise it is labeled as $1$. This particular choice was made according to the idea that we are more interested in detecting relevant changes at the cost of a higher fraction of false positives rather than losing important information (it is up to the analysis on ground to use post-processing algorithms and discard fake changes).

\section{Model architectures}
The design of the model architectures is highly correlated to the approach used for change detection. As a consequence, the models developed for semantic segmentation could not be also used to perform classification. In the semantic segmentation approach, a UNet-like architecture~\cite{unet} was designed, as it's proven to be a very efficient kind of DL algorithm to provide results with the required dimensionality. It is essentially composed by two symmetrical CNNs: the first part extracts features from the input data through a series of convolutions and pooling operations, while in the second part transpose convolutional layers are used together with the skip-connection technique to recover spatial information from the previous layers and use them at new abstraction levels. The final output is obtained by a convolutional layer with a $1\times1$ kernel and softmax activation. For the classification task, a traditional CNN was used: after several convolutional and pooling layers, a flattening operation is performed to reduce the dimensionality of the processed data to $1$, and a few dense layers with a sigmoid activation function are used before the $1$-neuron sigmoid output layer, which provides the overall score indicating whether the image pair contains a significative amount of changes.

\begin{figure}[b]
    \centering
    \includegraphics[width=.9\textwidth]{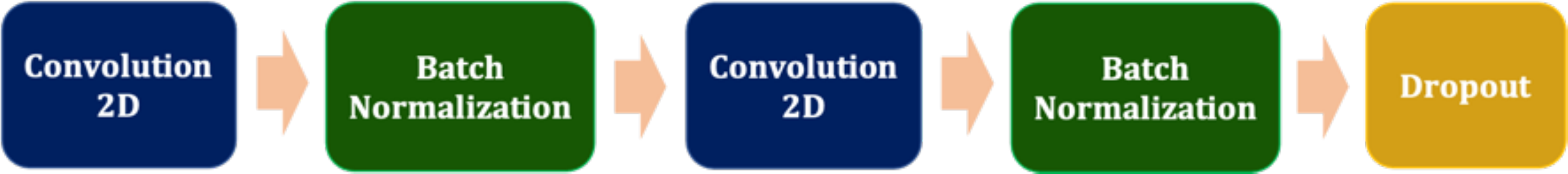}
    \caption{Basic convolutional unit used in all the model architectures.}
\label{fig:convunit}
\end{figure}

The main element of the neural networks that were designed for both the approaches is the convolutional unit. It has been shown that two or more consecutive convolutional layers with a small kernel (i.e. $3\times3)$ has the effect to enlarge the receptive field (the ``window'' from which features are extracted by a kernel) while keeping the number of parameters smaller with respect to using a single convolutional layer with a larger kernel~\cite{doubleconv}. Therefore, the basic convolutional unit used in all the models is composed by two stacked convolutional layers, each of them followed by a batch normalization layer that allows to obtain better convergence while speeding up the training process~\cite{bnorm}. Rectified linear unit (ReLU) was selected as the activation function of the convolutional layers, while a dropout layer~\cite{dropout} helps to reduce the overfitting effect and increase the model capacity. The logical structure of the convolutional unit is shown in Figure~\ref{fig:convunit}.

Another important step of the architecture design process was the choice of how the network is going to compare the two images in each pair to produce the corresponding output. Two possible methods were evaluated: the \textit{EarlyFusion} (EF) and the \textit{Siamese} (Siam) methods. In the EF method, the two images of a pair are concatenated along the channel dimension as the initial operation of the neural network; features are then extracted from a single image with size $128\times128\times26$. In the Siam method, instead, feature extraction is operated on each of the two images separately but using shared weights between the two paths created in the network, while the skip-connection steps recover the spatial information from both the processed images at different abstraction level.

\begin{figure}[tb]
    \centering
    \includegraphics[width=.55\textwidth, angle=-90]{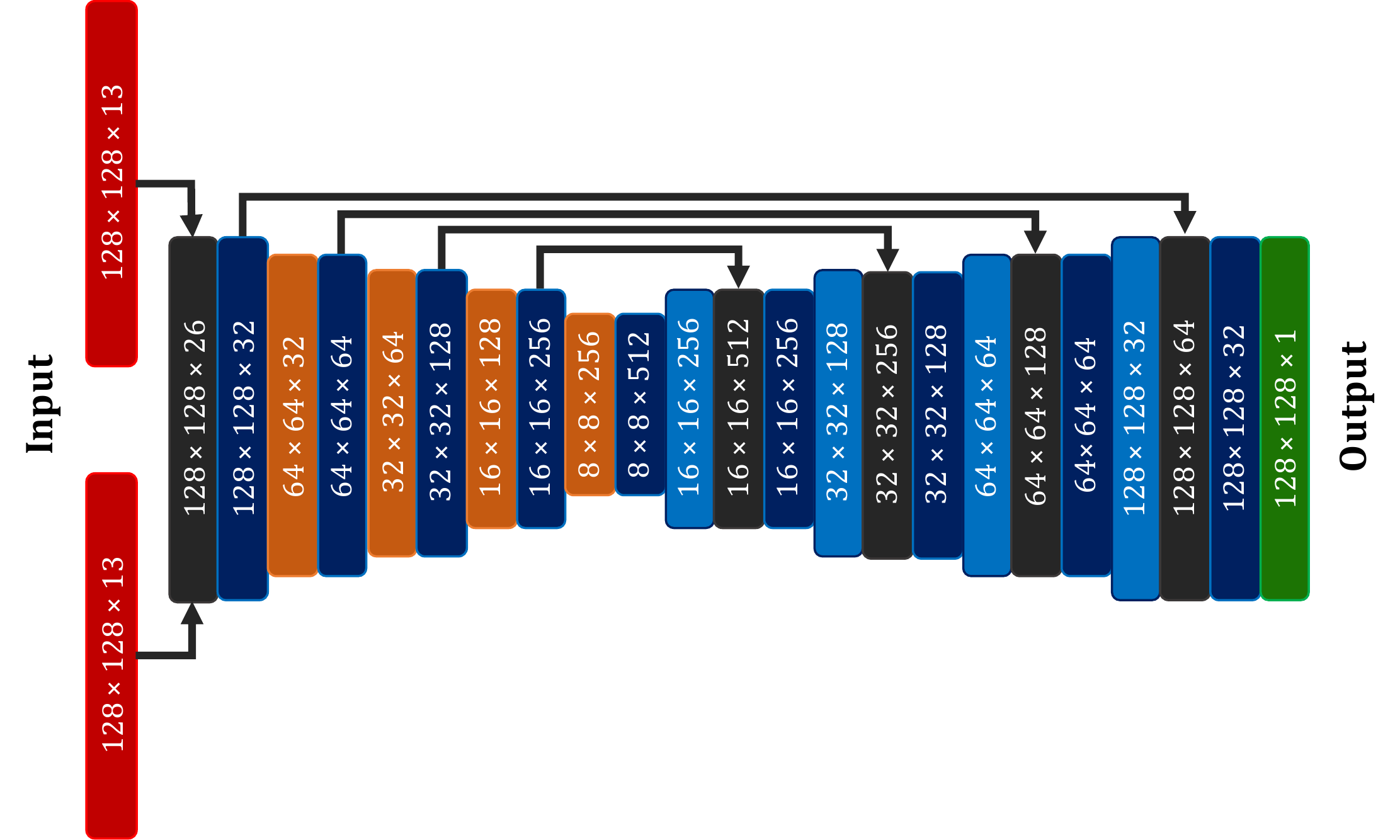}
    \hspace{2.5cm}
    \includegraphics[width=.55\textwidth, angle=-90]{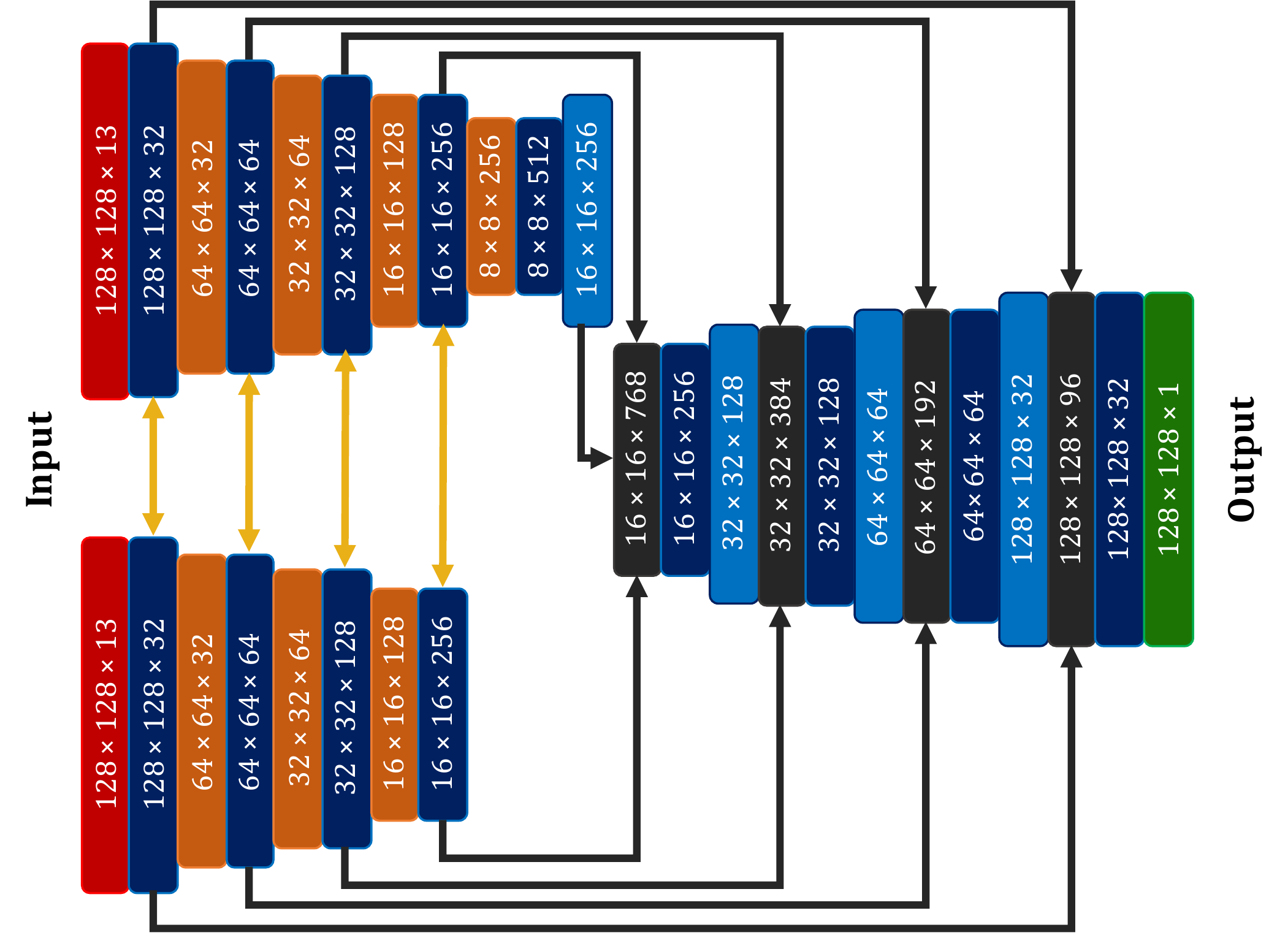} \\
    \includegraphics[width=.4\textwidth, angle=-90]{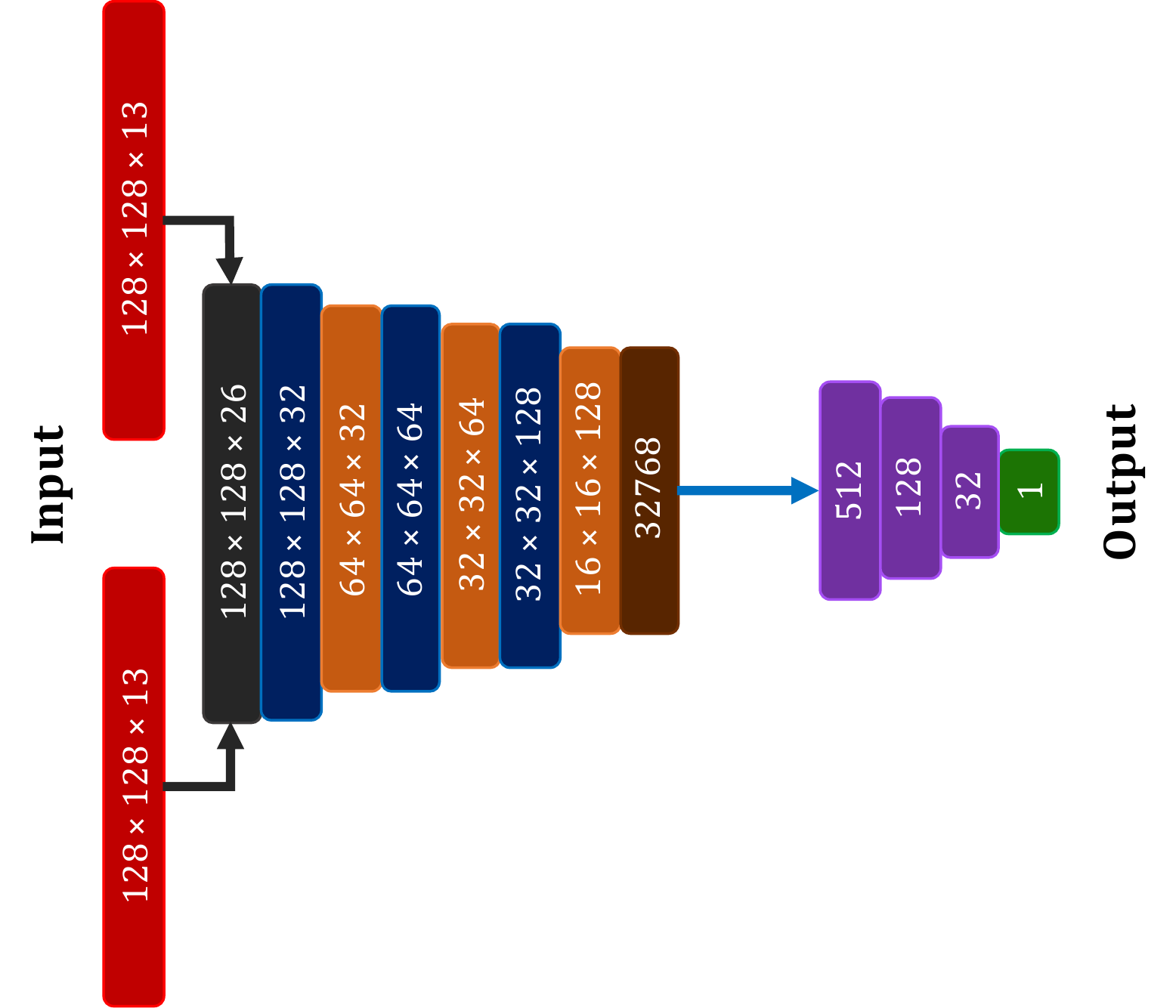}
    \caption{(\textit{Top}) Model architectures used in the change detection task with the segmentation approach. The model on the left adopts the \textit{EarlyFusion} method, while the model on the right uses the \textit{Siamese} method. (\textit{Bottom}) Model architecture used in the change detection task with the classification approach, adopting the \textit{EarlyFusion} method only. Numbers indicate the output size of each layer of the network. Red is for inputs, dark blue for convolutional units, orange for max pooling layers, light blue for transpose convolutional layers, dark gray for concatenation layers, brown for the flatten layer, violet for dense layers and green for output. Dark gray arrows indicate the concatenation operation, yellow arrows the shared weights and azure arrow is the batch normalization operation after flattening.}
\label{fig:architectures}
\end{figure}

Generally, the \textit{Siamese} method makes both the training and inference processes slower, because the concurrent feature extraction doubles the size of the model in the first part of the network, while this does not occur with the \textit{EarlyFusion} method. Thus, if $16$ different kernels are used in the first level of the neural network, $16$ feature maps are then produced from each of the two images, for a total of $32$ feature maps, whereas in the common feature extraction of EF only $16$ feature maps are produced in total. Moreover, a CNN cannot fully exploit the common feature extraction part of the Siam UNet, since the skip-connection technique is not used (the classification network does not recover the spatial information from previous layers). Therefore, since the classification approach mainly aims at achieving very fast execution in the onboard scenario, only the EF method was adopted for this task. The final architectures for classification and semantic segmentation are shown in Figures~\ref{fig:architectures}.

\section{Training details}
The training process was designed having in mind that the dataset is poor and extremely unbalanced with respect to the class population; therefore, the choice of the loss function was crucial.

The unbalance problem is a distinctive feature of change detection tasks. Indeed, changes can be considered rare events with respect to the pixel content of a satellite image, especially with a resolution of $10$~m, where most of the changes involve very small areas (of the order of few pixels). In the case of the OSCD dataset, the ratio between the number of changed pixels and the total number of pixels is extremely small (about $2\%$); therefore, any classifier trained without using methods to handle the unbalance between class samples would not be able to recognize changed pixels and most probably classify every pixel as unchanged, achieving apparently high performance.

Two solutions exist to solve this problem: a) balancing the dataset by removing some samples from the most populated class and having more or less the same number of samples for both the classes, and b) assigning weights to samples depending on their class population, during the minimization process of the loss function. The latter solution was adopted because the former method is not suitable for change detection. Indeed, the model needs to be trained also on image pairs which do not contain any change, otherwise the algorithm will be biased and find at least the $2\%$ of changed pixels within new scenes.

Two loss functions are used in this study. The first, denoted with ``bce'', is the binary cross-entropy with the introduction of weights that are inversely proportional to the class population. For a batch of $N$ image pairs extracted from the training dataset, each having $M$ pixels, the loss function to minimize at each iteration is
\begin{equation}
\label{eqn:lossbce}
    \mathcal{L}^{bce} = -\frac{1}{N\times M}\sum_{i=1}^{N\times M}{w_0t_{0i}\log{s_{0i}} +w_1(1-t_{0i})\log{(1-s_{0i})}}
\end{equation}
where $i$ denotes the pixel, $w_0$ and $w_1$ are the weights for the two different classes, $t_0$ and $t_1$ are the target labels at the two classes, while $s_0$ and $s_1$ are the predictions for the two classes. Both the target labels and predictions are used in the form of one-hot vectors (arrays in which each element corresponds to the probability of the sample to belong to each class).

The second loss function tested, denoted with ``wbced'', is the sum of the loss function of Eq.~\ref{eqn:lossbce} and the dice function
\begin{equation}
\label{eqn:dice}
    \mathcal{L}^{dice} = 1 - \frac{2y_p y_t+1}{y_p+y_t+1}
\end{equation}
where $y_p$ is the predicted class and $y_t$ the ground truth, evaluated with binary method ($1$ for change, $0$ otherwise). Dice loss is often adopted for segmentation tasks, as the dice coefficient (the fraction of Eq.~\ref{eqn:dice}) is a helpful metric to evaluate the similarity of contour regions~\cite{dice} and to take into account the distribution of changed pixels across the change map, which is not uniform. This loss function was inspired by the study made with the UNet++ model for very-high resolution satellite images~\cite{unet++}.

While both the considered loss functions are used for the segmentation approach, only the bce loss function is adopted for the classification task (without summing and averaging over the number of pixels $M$). In addition, the k-fold cross validation strategy is adopted in both the approaches, as it represents an efficient method to evaluate models' performances when dealing with poor statistics datasets. In this study, k is set equal to $5$.

\section{Performance study}
Several metrics were considered to evaluate the performance of each model. Given that the change detection task is a binary classification problem (either pixel-based or patch-based), the changed samples are considered ``positives'', as they carry the information of interest, while the unchanged samples are considered ``negatives''. The following notation is then adopted: $T_P$ for true positives, $T_N$ for true negatives, $F_P$ for false positives and $F_N$ for false negatives.

The classical accuracy score (fraction of correctly classified samples with respect to the entire set of data available, independently of the class population) is not suitable for the purpose, as it would overestimate the goodness of the predictions due to the unbalance problem. Indeed, if the $98\%$ of the data is labeled as unchanged, any classifier could get $98\%$ accuracy while predicting each sample as belonging to that class (included the changed samples), thus committing an error of $2\%$ and having no capability to detect any change at all. Nevertheless, a \textit{balanced accuracy} can be defined as
\begin{equation}
\label{eqn:bacc}
    balanced\;accuracy = \frac{1}{2}\left(\frac{T_P}{T_P+F_N}+ \frac{T_N}{T_N+F_P}\right)
\end{equation}
in order to represent the mean value of the \textit{accuracy} scores evaluated on the two classes separately. Although \textit{balanced accuracy} is a good metric to evaluate model performance for tasks with unbalanced datasets, in this study three additional scores were considered. The \textit{precision}
\begin{equation}
\label{eqn:prec}
    precision = \frac{T_P}{T_P+F_P}
\end{equation}
represents the fraction of positive predictions that are actually correct. The \textit{recall} or \textit{sensitivity}
\begin{equation}
\label{eqn:rec}
    recall = \frac{T_P}{T_P+F_N}
\end{equation}
represents the fraction of true positives that are correctly identified. Finally, the \textit{F1-score}
\begin{equation}
\label{eqn:f1}
    F1 = 2\times \frac{precision \times recall}{precision+recall}
\end{equation}
combines \textit{precision} and \textit{recall} to provide a new reliable metric.

In this study, the \textit{recall} was selected as the target score to maximize. Indeed, it is crucial to reduce the fraction of false negatives, namely those changes that are wrongly predicted by a certain model and thus lost, while accepting a higher number of false positives that can be later rejected by additional post-processing algorithms. The other metrics were also considered to ensure that the models have reliable performance despite the choice of maximizing the \textit{recall}.
 
In the following section the results for the segmentation approach are presented, whereas those for the classification approach are discussed in the subsequent one.

\subsection{Segmentation results}
As previously mentioned, two different architectures are studied (EarlyFusion and Siamese, see Figure~\ref{fig:architectures}) and two loss functions are tested during the training process. As a result, four different segmentation models are studied and denoted with ``Architecture-loss'' logic: EF-bce, Siam-bce, EF-wbced and Siam-wbced.

\begin{figure}[!hb]
    \centering
    \includegraphics[width=.4\textwidth]{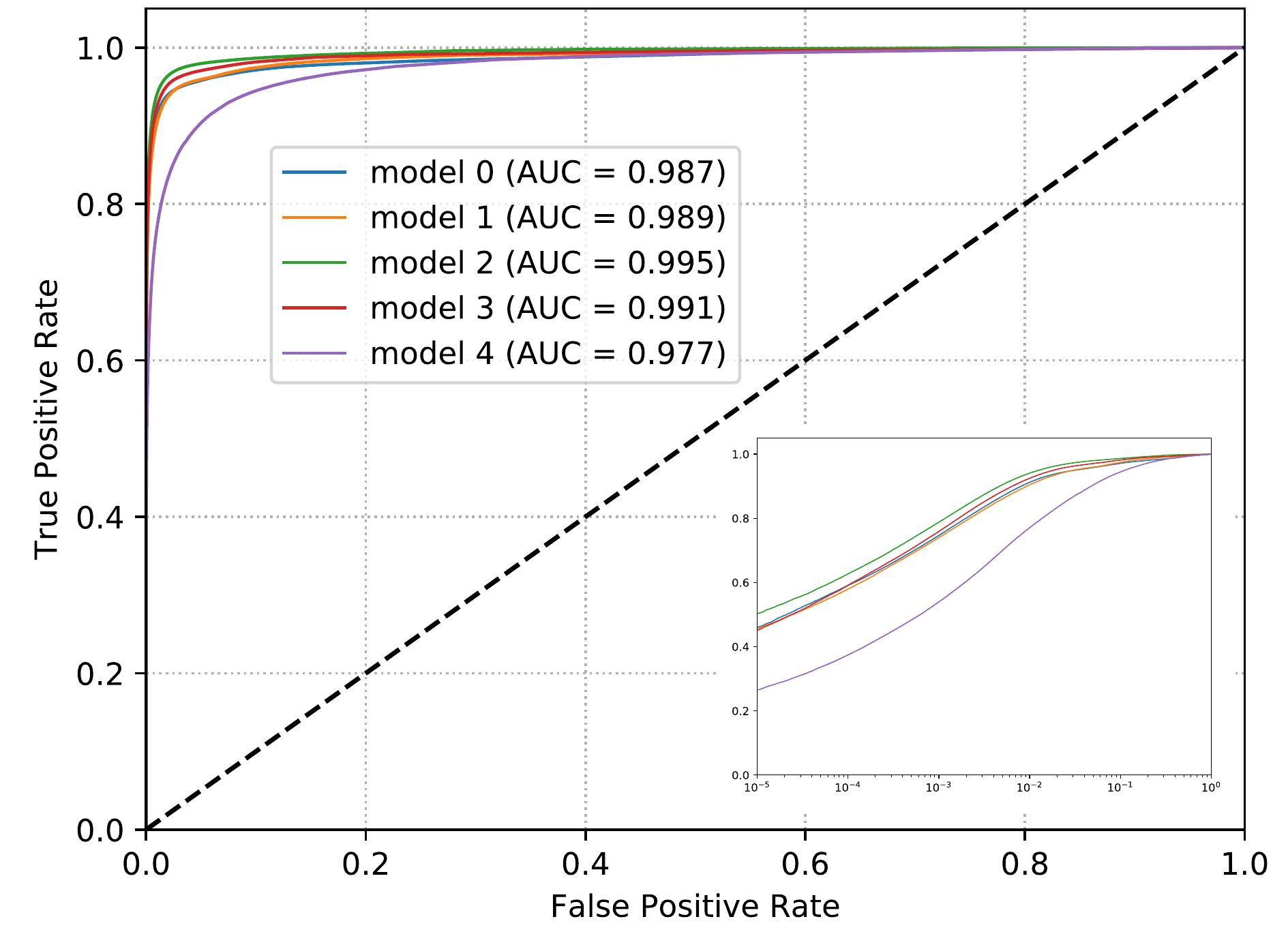}
    \hspace{0.4cm}
    \includegraphics[width=.4\textwidth]{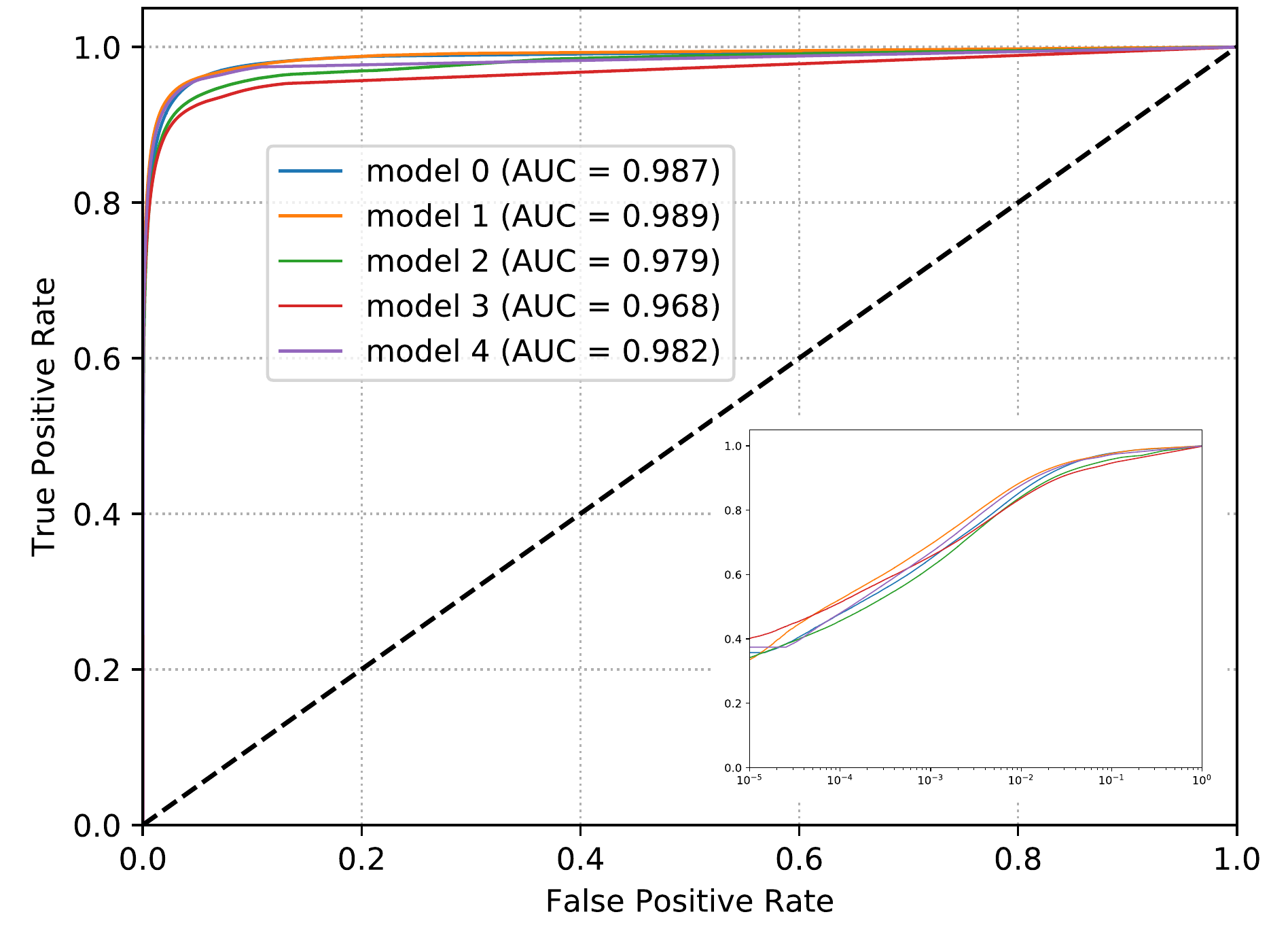} \\
    \vspace{0.4cm}
    \includegraphics[width=.4\textwidth]{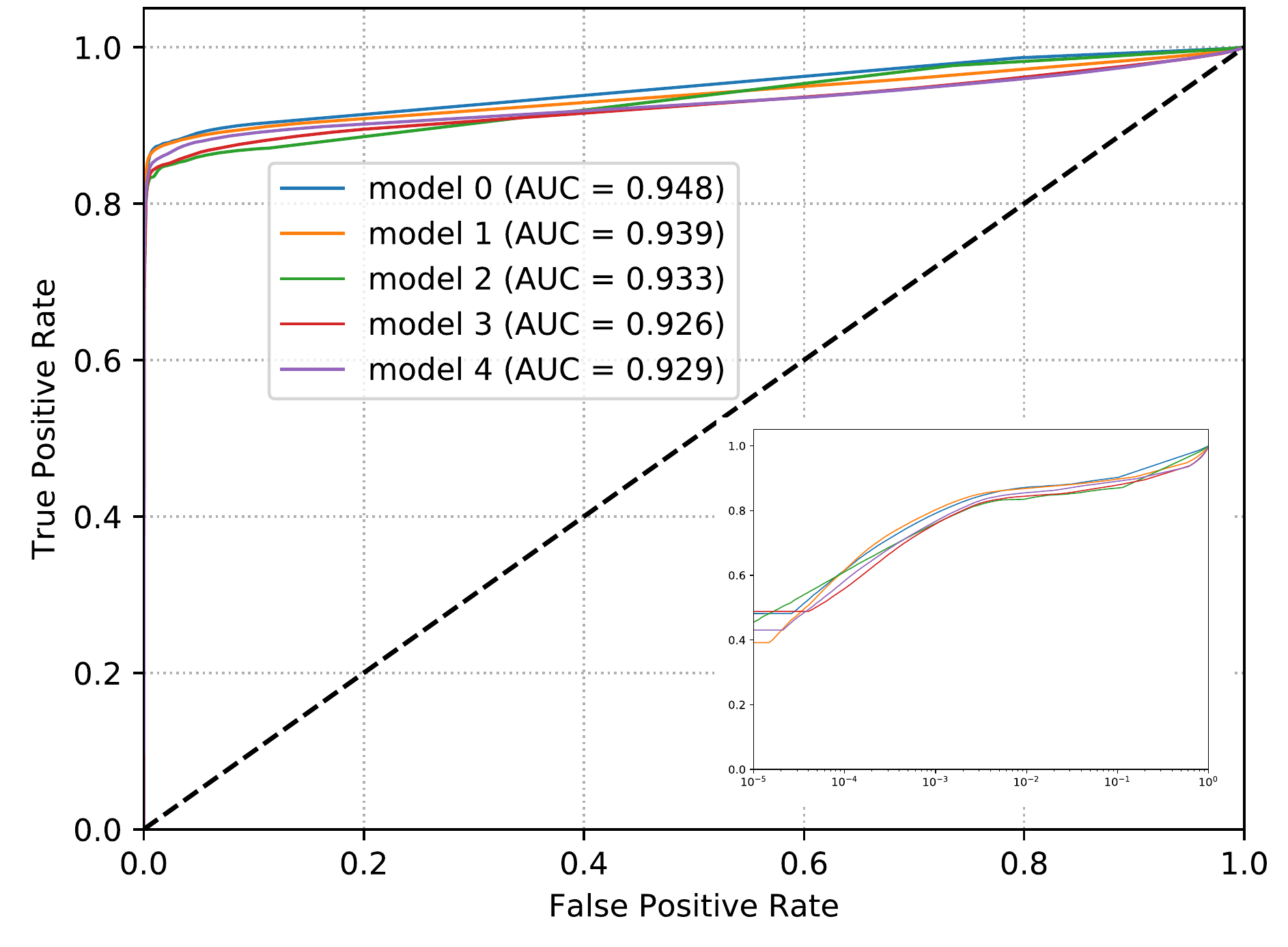}
    \hspace{0.4cm}
    \includegraphics[width=.4\textwidth]{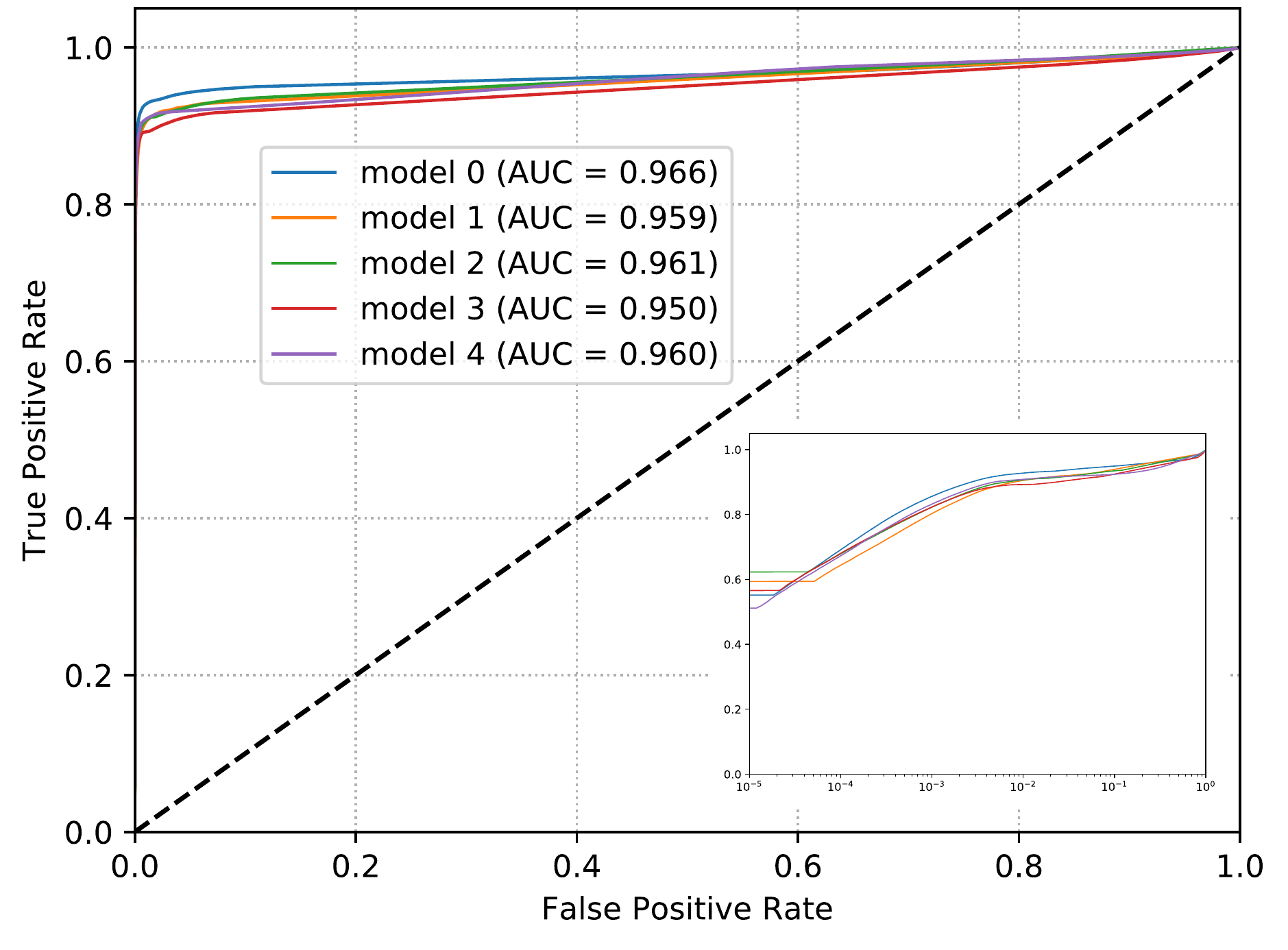}
    \caption{ROC curves of the five models trained with the cross validation technique for the segmentation task. Top are EF-bce (\textit{left}) and Siam-bce (\textit{right}) models, while bottom are EF-wbced (\textit{left}) and Siam-wbced (\textit{right}) models. The inset plots show the ROC curves in logarithm scale for the $x$-axis.}
\label{fig:roc}
\end{figure}

\newpage

\begin{table}[!ht]
\centering
\resizebox{.9\textwidth}{!}{%
\begin{tabular}{c|c|c|c|c}
\textbf{Model} & \textbf{Precision} & \textbf{Recall} & \textbf{F1} & \textbf{Balanced Accuracy} \\
\hline
\textit{EF-bce}     & $91.45 \pm 1.68$ & $75.26 \pm 8.71$ & $82.33 \pm 5.96$ & $87.55 \pm 4.35$ \\ \hline
\textit{Siam-bce}   & $86.73 \pm 3.71$ & $73.93 \pm 4.98$ & $79.60 \pm 1.94$ & $86.83 \pm 2.45$ \\ \hline
\textit{EF-wbced}   & $94.22 \pm 0.85$ & $78.24 \pm 1.72$ & $85.47 \pm 1.06$ & $89.06 \pm 0.86$ \\ \hline
\textit{Siam-wbced} & $95.50 \pm 0.60$ & $82.14 \pm 1.94$ & $88.30 \pm 1.04$ & $91.03 \pm 0.97$ \\\hline
\end{tabular}%
}
\vspace{3mm}
\caption{Average percentage values of the considered metrics for the segmentation task. The uncertainties are estimated as the standard deviation over the five models trained with the k-fold cross validation.}
\label{tab:seg_scores}
\end{table}
The ROC curves are shown in Figure~\ref{fig:roc}. Each plot presents the ROC curves obtained by the five models trained with the k-fold cross validation technique, for each architecture and loss function. These results show that models trained with the bce loss function achieve higher \textit{AUC} (\textit{Area Under Curve}) scores than models trained with wbced, despite the latter loss function takes into account additional features of the change detection problem (i.e. clusters of changed pixels). 
The ROC curves prove that the trained models have good overall performance and are useful metrics to choose the working point for a certain task. In this study, however, they were used to observe the trend of the true positive rate (or \textit{recall}) with respect to the false positive rate across the five models. Specifically, anomalous performance is observed for model~4 of EF-bce (top left), where the \textit{AUC} score is lower than the other models of the same sample thus indicating that a possible issue occurred during training, such as the loss function stuck in a local minimum or a problem related to the dataset partition. Since this effect is not present for other architectures and loss functions, where the same partitions are used, the second hypothesis was discarded. This issue can be thus considered a statistical effect due to the training process and not a specific feature of the models or task configuration.

The choice of the working point for each model was operated taking into account the average scores across the five trained model for different change thresholds. Indeed, the outputs of the segmentation models are change maps with each pixel having a value that ranges from $0$ to $1$. Setting a change threshold at $0.5$ means that all the pixels above such value will be rounded to $1$, or to $0$ otherwise, but it does not necessarily represent the optimal choice for this study. As previously mentioned, the target performance to achieve in this work was to maximize the \textit{recall} metric, while checking the other scores to verify the reliability of the evaluated model performance. The corresponding change threshold was identified as the value of $0.3$, and was adopted for all the segmentation models whose scores and results are presented in this section. 

Table~\ref{tab:seg_scores} shows the average scores of the four segmentation algorithms and their uncertainties, calculated as the standard deviations across the five models trained with the k-fold cross validation technique. Models trained with the wbced loss function seem to have better overall performance, achieving a \textit{recall} of $78.24\%$ (EF-wbced) and $82.14\%$ (Siam-wbced) and a \textit{F1-score} larger than $85\%$. Furthermore, errors are smaller than $1\%$, thus indicating more stability during the training process with respect to the dataset partitions. On the other hand, models trained with the bce loss function have larger errors and lower \textit{recall} (respectively $75.26\%$ and $73.93\%$ for EF-bce and Siam-bce), but still be able to achieve promising results.

Combining the observations made on the scores with the ROC curves seen in Figure~\ref{fig:roc}, it is natural to suspect that a larger overfitting effect characterizes the training process for models where the wbced loss function is used. Indeed, such loss function was originally designed to work with challenging very-high resolution images, and might not be optimal for Sentinel-2 images. At the same time, using a loss function that is already integrated in almost every deep learning framework with the addition of weights inversely proportional to classes populations seems to provide very good results and can be a choice that better meets the target of this study.

\begin{figure}[ht!]
    \centering
    \includegraphics[width=.35\textwidth]{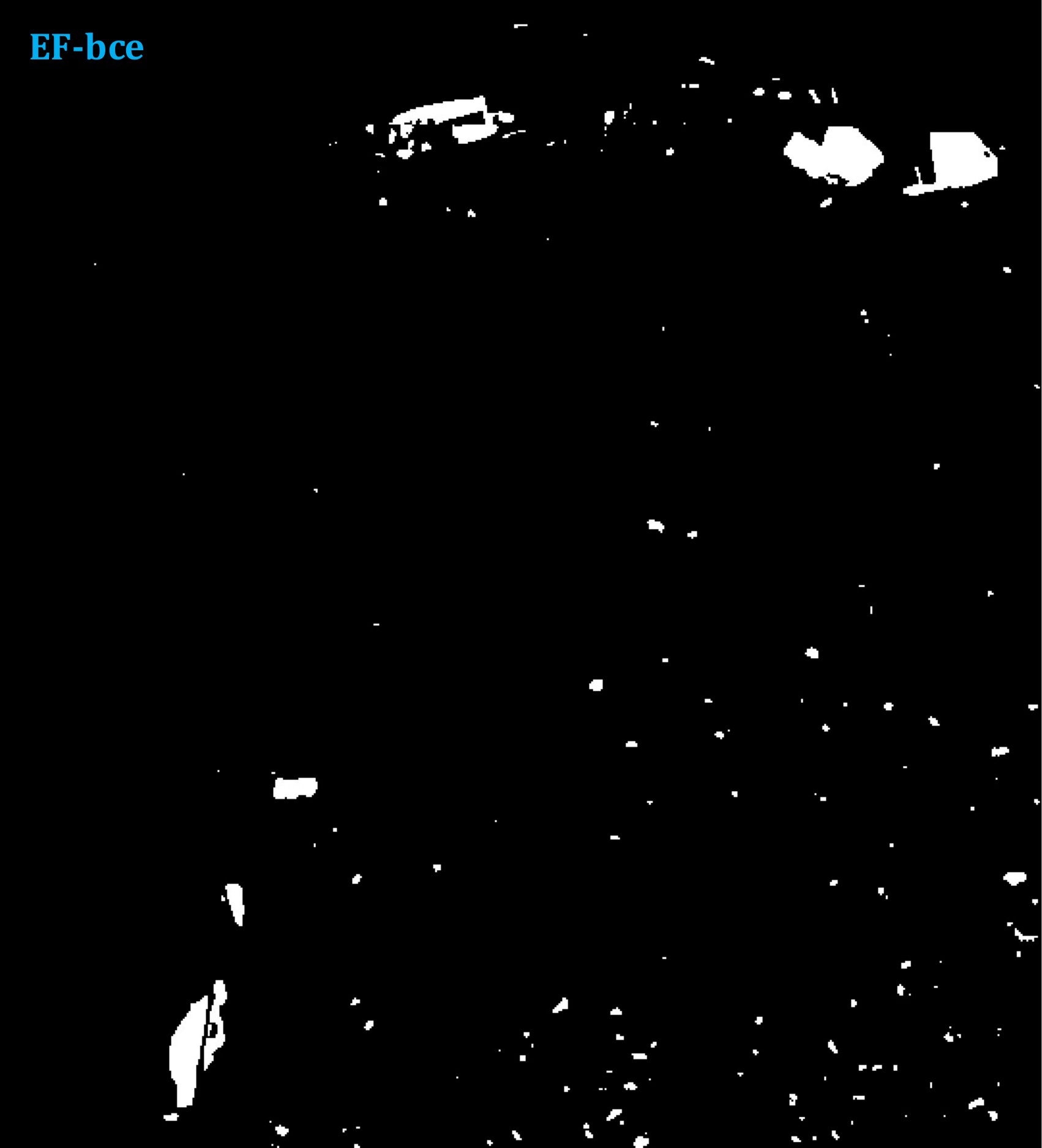}
    \hspace{0.5cm}
    \includegraphics[width=.35\textwidth]{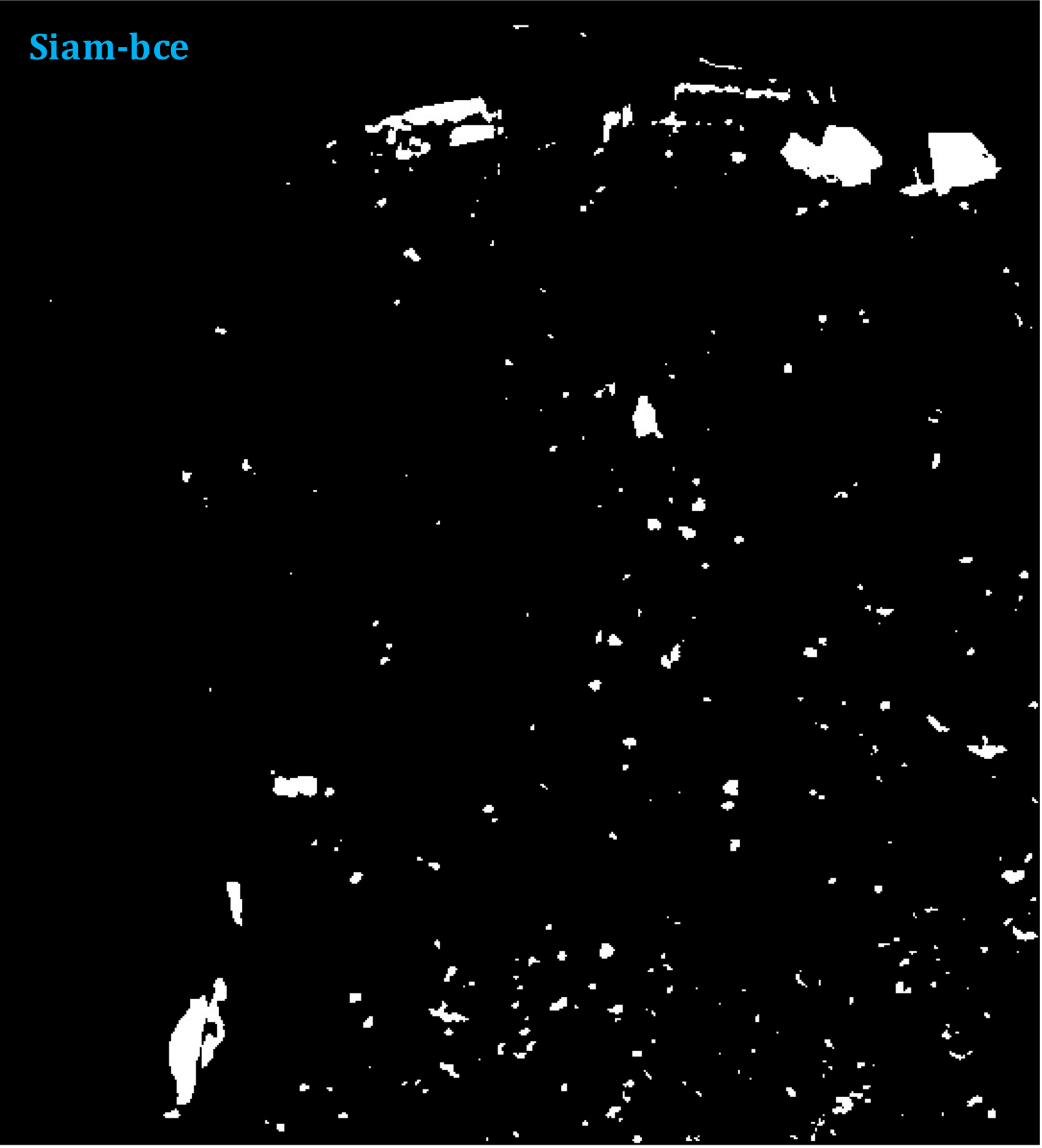} \\
    \vspace{0.5cm}
    \includegraphics[width=.35\textwidth]{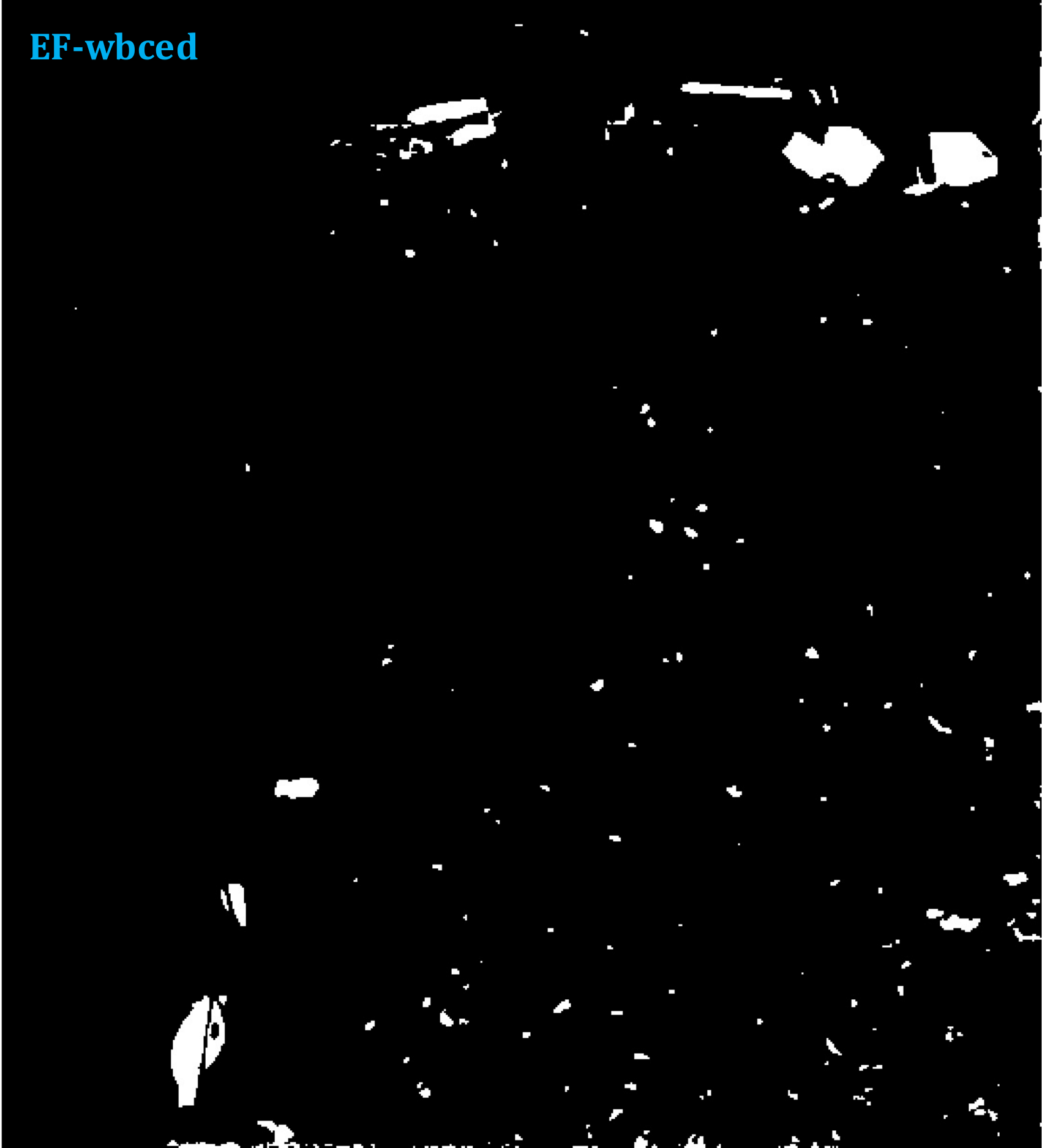}
    \hspace{0.5cm}
    \includegraphics[width=.35\textwidth]{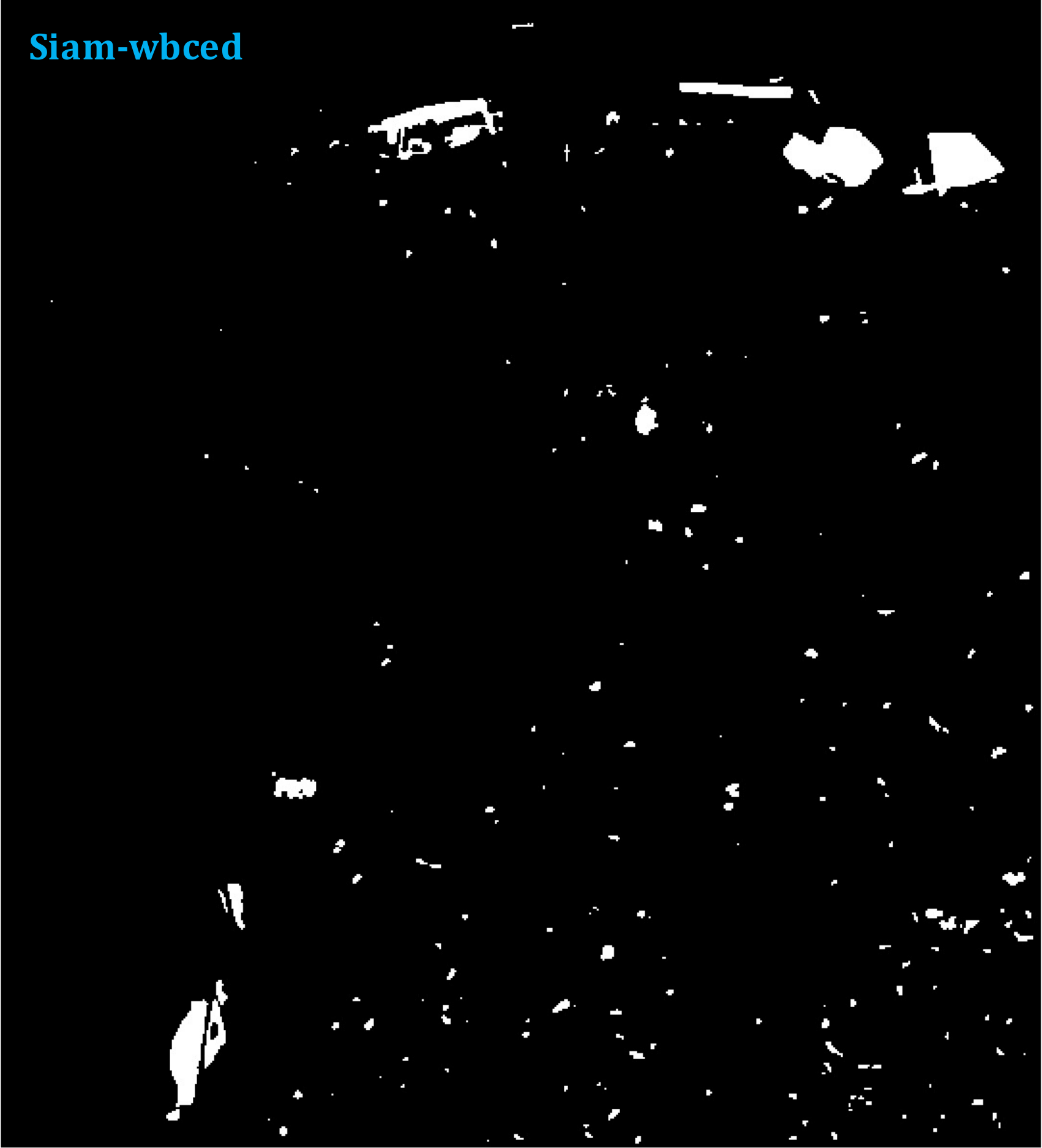}
    \caption{Change maps of the Beirut city obtained by the four segmentation models: EF-bce (\textit{top left}), Siam-bce (\textit{top right}), EF-wbced (\textit{bottom left}) and Siam-wbced (\textit{bottom right}).}
\label{fig:beirut-cm}
\end{figure}

Figure~\ref{fig:beirut-cm} shows the change maps obtained by the four segmentation models for the Beirut city, seen in Figure~\ref{fig:beirut_sample}. Change maps obtained by EF-wbced and Siam-wbced confirm the suspect above mentioned about a larger overfitting effect occurring with these models. Indeed, some clusters of changed pixels are detected with a pixel-wise precision, while other small changes are not detected at all. A more ``realistic'' result is provided by the model EF-bce and Siam-bce, where changes are detected at the correct positions, but a refinement operation is needed in a post-processing phase. The advantage, in this case, is that they can provide better overall results on test data. Furthermore it is important to remark that the average scores are calculated during the model validation phase and are based on the scores obtained on dataset partitions that still belong to the same original dataset, even though not used at training time.

The goal of the segmentation task was to study the performance of different architectures while working with the same dataset, and observe how a more complex loss function might improve or worsen the results. A test was performed with EF-bce on a completely different dataset, and presented in Section~\ref{sec:rome}.

\newpage

\begin{figure}[hb!]
    \centering
    \includegraphics[width=.45\textwidth]{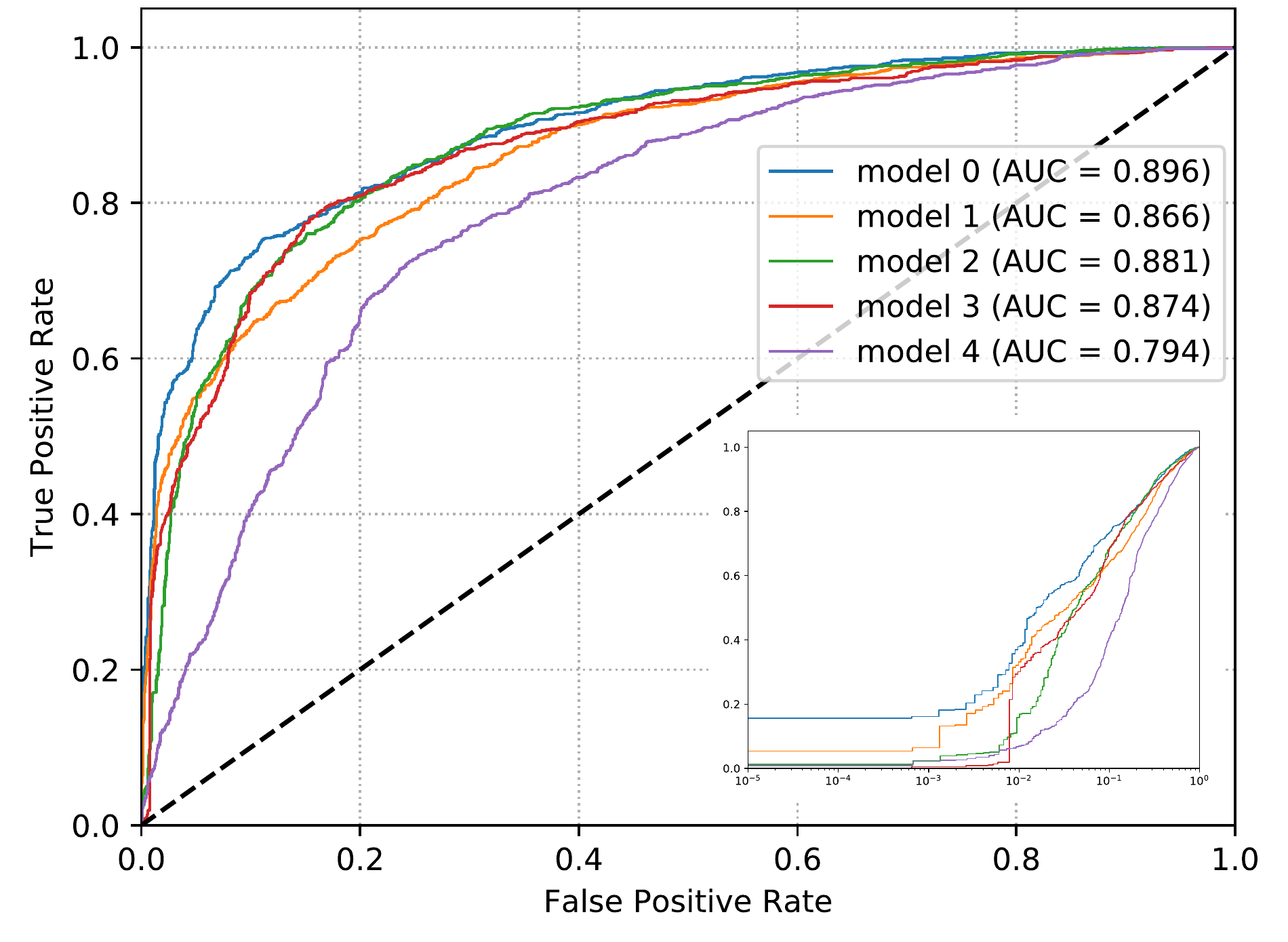}
    \caption{ROC curves of the five models trained with the cross validation technique for the classification task. The inset plot shows the ROC curves in logarithm scale for the $x$-axis.}
\label{fig:roc_class}
\end{figure}

\newpage
\begin{table}[ht!]
\centering
\resizebox{.9\textwidth}{!}{%
\begin{tabular}{c|c|c|c|c}
\textbf{Model} & \textbf{Precision} & \textbf{Recall} & \textbf{F1} & \textbf{Balanced Accuracy} \\
\hline
\textit{EF-CNN}     & $73.98 \pm 5.73$ & $80.16 \pm 2.77$ & $76.72 \pm 2.44$ & $77.91 \pm 3.43$ \\ \hline
\end{tabular}%
}
\vspace{3mm}
\caption{Average percentage values of the considered metrics for the classification task. Uncertainties are calculated as the standard deviation over the five models trained with the k-fold cross validation.}
\label{tab:class_scores}
\end{table}

\subsection{Classification results}
As previously mentioned, only the EarlyFusion method was used in the design of the classification model to compare the two images before and after changes. The developed model was named ``EF-CNN''.

Figure~\ref{fig:roc_class} shows the ROC curves of the five models trained with k-fold cross validation. In this case, \textit{AUC} scores are lower with respect to the segmentation task as expected. Indeed, the definition of ground truth might be ambiguous for certain samples, because the image pair was labeled as ``changed'' if at least $25$ pixels were actually changed pixels, but no restrictions were set with respect to their distribution across the entire scene (i.e. presence of a large cluster, or changes covering a few pixels each, etc.). Nevertheless, results are very promising (\textit{AUC} scores still above $85\%$ with the exception of model~4 that might feature a training issue as experienced in the segmentation approach) if considering that the target workflow of this study is to detect possible changes onboard the satellite and, in the positive case, transmit the new acquired image to the ground stations.

In the classification approach, the change threshold  was also set to $0.3$. In this case, such value is related to the entire image pair and not considered for single pixels, and is used to establish if the pair contains a considerable number of changes along the period of observation or not. 

Table~\ref{tab:class_scores} shows the average scores of the classification model and the corresponding uncertainties, calculated as the standard deviations across the five models trained with the k-fold cross validation. Despite the ambiguous definition of the ground truth as previously mentioned, the \textit{recall} is still around the $80\%$ and the \textit{F1-score} is around $77\%$, which represents a good result achieved by this enough simple deep learning algorithm. 

\begin{figure}[b]
    \centering
    \includegraphics[width=.5\textwidth]{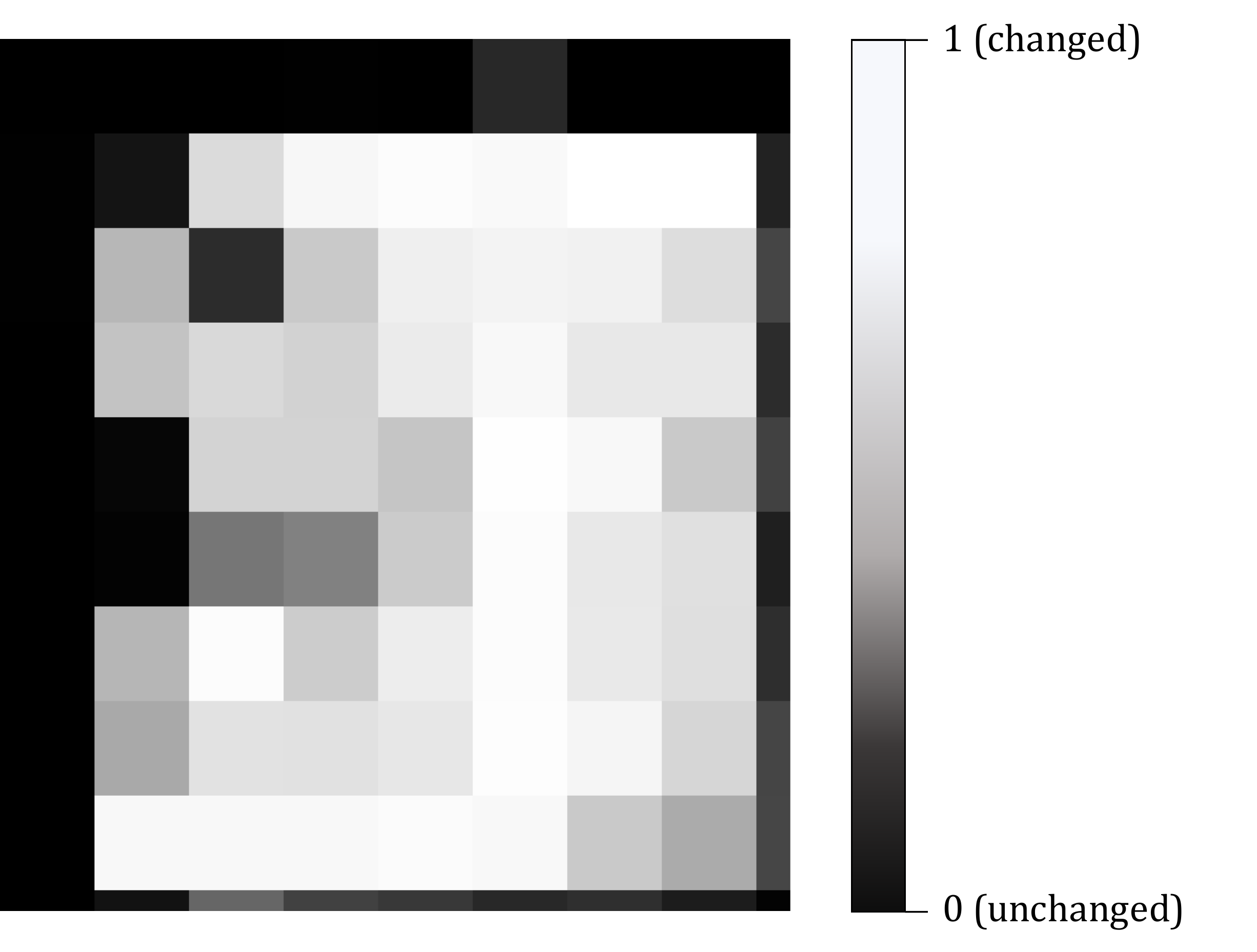}
    \caption{Coarse change map of the Beirut city obtained by the classification model EF-CNN. The darker the section of the map, the lower the probability of having a considerable amount of changes in the area.}
\label{fig:beirut-greycm}
\end{figure}

An alternative output of the classification model, if not using the change threshold, is a coarse version of the change map. All the pixels of each $128\times128$ patch are assigned the output of the CNN for the corresponding image pair, and greyscale is used while combining all the patches back to form the final change map: the darker the patch, the lower the probability of having a considerable amount of changes in that section of the scene. The greyscale change map for the Beirut area is shown in Figure~\ref{fig:beirut-greycm}. It can be observed how the left border of the map is completely black, as no changes are present in that part of the scene; the $128\times128$~patches that cover a large cluster of changed pixels are white, while intermediate situations are generally represented with different shades of grey. This result is generally useful to study how the algorithm is performing on the entire scene, while in the onboard scenario only the CNN output is needed, as it represents the discriminating score that permits to decide onboard whether to transmit the patch to the ground station.

\newpage
Finally, it is important to remark that the classification study is performed assuming that the onboard pre-processing phase includes the operation of co-registration of images acquired on the same region at different times, that represents a possible scenario of future satellite missions. If the image co-registration is not guaranteed onboard, the trained algorithm would have worse performance since the change detection algorithms are usually very sensitive to this kind of related issues.

\section{Test case: Rome Fiumicino airport}
\label{sec:rome}

The EF-bce model was tested on a new dataset covering another area of interest namely the Rome Fiumicino airport. The ground truth manually detected by photointerpretation for this area is only available for very-high resolution images and, therefore, not all of the labeled changes could be detected on the corresponding Sentinel-2 images, depending on their size. Furthermore, not all of the changes present in the ground truth are related to urban variations (the dataset was created for different target studies), while at the same time some urban changes are not labeled at all. The period of observation is December 2018 - September 2019.

\begin{figure}[!b]
    \centering
    \includegraphics[width=.9\textwidth]{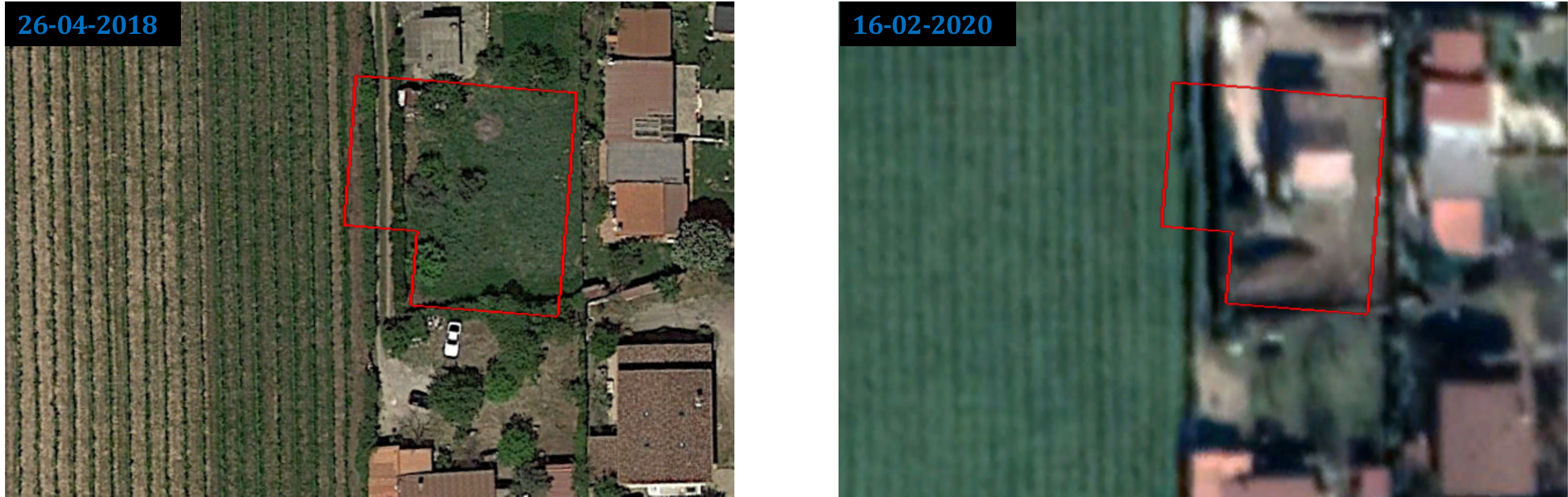}
    \caption{Scene comparison of the Rome Fiumicino airport area. The image on the left has very-high resolution, while the image on the right is acquired by the Sentinel-2 satellite mission. The algorithm correctly detects a new house built in the scene.}
\label{fig:rome1}
\end{figure}

In order to apply the developed change detection algorithm to this dataset, it was necessary to select only those changes covering an area of $1600$~m$^2$ or larger in the ground truths (corresponding to an area of $16$ pixels in Sentinel-2 images). Then, after producing the change map with the EF-bce model, a filtering operation was performed by cross-checking the results with available CORINE Land Cover~\footnote{https://land.copernicus.eu/pan-european/corine-land-cover} maps and Google Earth\footnote{https://www.google.com/earth/} images to discard respectively changes in non-urban areas and those along the coast, which could be caused by a larger error in the image co-registration.

A total of $38$~changes were detected as passing this filtering phase. While $24$ of these changes were confirmed also by human eye, the remaining $14$ are $F_P$ and considered particular cases (shadow effects, vegetation changes in the airport area, presence of airplanes along the runways, etc.). The presence of such a relatively high number of false positives is compatible with the initial decision of detecting as many changes as possible and eventually discard them in the post-processing phase, in order to maximize the efficiency of the change detection.

\begin{figure}[!t]
    \centering
    \includegraphics[width=.9\textwidth]{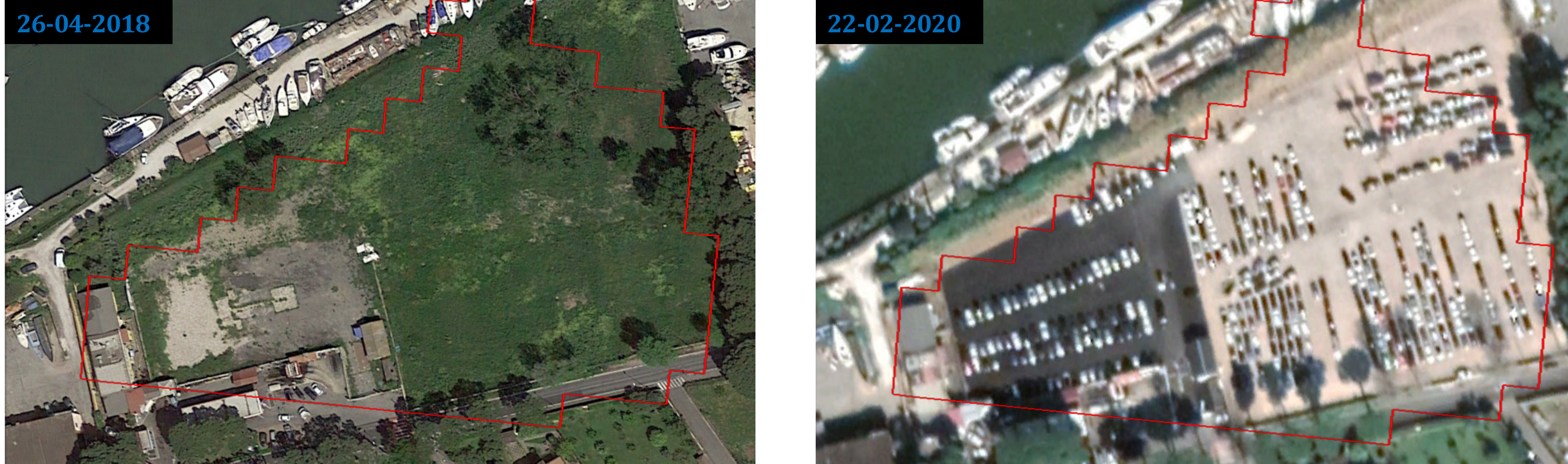}
    \caption{Scene comparison of the Rome Fiumicino airport area. The image on the left has very-high resolution, while the image on the right is acquired by the Sentinel-2 satellite mission. The algorithm correctly detects a new parking area in the scene.}
\label{fig:rome2}
\end{figure}

Finally, the $14$~changes of the available ground truth that passed the filtering selection about the pixel size were intersected with the changes detected by the algorithm; $5$ of them were correctly detected, while $9$ are $F_N$. The final results thus consist of $33$~true changes, with $24$ $T_P$, $9$ $F_N$ and $14$ $F_P$, for an overall \textit{recall} score of $72.73\%$ and a \textit{precision} value of $63.16\%$. These results confirm the performance obtained with the OSCD dataset and show that the post-processing phase has great importance in change detection studies to enhance the efficiency of the algorithms. Some of the change detection results are shown in Figure~\ref{fig:rome1} and~\ref{fig:rome2}. The images on the left have very-high resolution (the same for the available ground truth), while those on the right are acquired by the Sentinel-2 satellite mission. The red box outlines the changed area detected by the algorithm.

\section{Conclusions}
The work presented in this paper shows very promising results, that can be considered as a baseline for future studies. Two different architectures were developed for the segmentation approach, while also testing two loss functions: a ``classical'' binary cross-entropy and its sum with the dice function. In both cases, it was crucial to apply weights in order to face the unbalance problem typical of the change detection datasets. Although the performance of the models is similar, those models trained with the simple binary cross-entropy loss experience a smaller overfitting effect in the case of Sentinel-2 images and should be preferred, thus limiting the choice of the other loss function to the case of very-high resolution images, where contour regions are sharply defined. For the classification task, instead, a very simple CNN model was developed, with promising results that might be improved with a more robust definition of ground truth. Finally, one of the segmentation models was tested on a new observed area, where its performance matched the results obtained with the validation dataset within an expected small loss of performance typically occurring with new data.

\section*{Acknowledgements}
The authors would like to thank the ReCaS\footnote{https://www.recas-bari.it/index.php/en/} (a project funded by the Italian Ministry for Education, University and Research in the PON Research and Competitiveness 2007-2013 Notice 254 / Ric) data center management team, that provided access to powerful and high-performance devices to support the work here presented.

\bibliographystyle{unsrt}
\bibliography{references} 

\end{document}